\documentclass[conference]{IEEEtran}
\IEEEoverridecommandlockouts

\pdfoutput=1

\usepackage{cite}
\usepackage{amsmath}
\usepackage{amssymb}
\usepackage{booktabs}
\usepackage{graphicx}
\usepackage{multirow}
\usepackage{url}
\usepackage[hidelinks]{hyperref}
\usepackage{algorithm}
\usepackage{algorithmic}
\usepackage{placeins}
\usepackage[T1]{fontenc}
\usepackage[utf8]{inputenc}

\newcommand{\Description}[1]{}

\AtBeginDocument{%
  \setlength{\abovedisplayskip}{2pt plus 1pt minus 1pt}%
  \setlength{\belowdisplayskip}{2pt plus 1pt minus 1pt}%
  \setlength{\abovedisplayshortskip}{1pt plus 1pt}%
  \setlength{\belowdisplayshortskip}{2pt plus 1pt minus 1pt}%
}
\title{\LARGE \bf Underwater Visual Target Tracking with Target-Specific Depth Estimation and Adaptive Model-Fusion Predictive Control}

\author{Yuheng Zhou$^{1}$, Haiyang Cheng$^{1}$, Yanqi Feng$^{1}$, Pangkit Fong$^{1}$,\\
Mei Xuan Lee$^{2}$, Marcus Gee$^{3}$, Chongrong Fang$^{1}$ and Jianping He$^{1*}$%
\thanks{$^{1}$Yuheng Zhou, Haiyang Cheng, Yanqi Feng, Pangkit Fong, Chongrong Fang and Jianping He are with the Department of Automation, Shanghai Jiao Tong University, Shanghai, China
{\tt\small \{wsklxzm6, chy\_maer, fengyanqi, fpjgaoge, crfang, jphe\}@sjtu.edu.cn}}%
\thanks{$^{2}$Mei Xuan Lee is with the University of Toronto, Toronto, Canada
{\tt\small austinmx.lee@mail.utoronto.ca}}%
\thanks{$^{3}$Marcus Gee is with the University of Calgary, Calgary, Canada
{\tt\small marcus.gee@ucalgary.ca}}%
}

\begin{document}

\maketitle
\thispagestyle{empty}
\pagestyle{empty}

\begin{abstract}
Vision-based underwater target tracking is challenged by unreliable depth measurements and unknown target motion. This paper proposes a stereo visual-servoing framework for an autonomous underwater vehicle (AUV). For perception, the framework derives a stable 3D relative state from stereo images through target-specific depth extraction and Kalman filtering. It constructs a target-depth mask from color, disparity, and temporal cues to select reliable target pixels, and then filters the resulting depth measurement and detected image center separately. For control, the framework decouples yaw regulation from translational control, avoiding computationally expensive coupled multi-DOF optimization and enabling real-time translational MPC. The translational controller employs adaptive model-fusion predictive control, combining constant-velocity and zero-velocity target models to accommodate different target-motion patterns. It updates the model weights using historical prediction errors and computes translational commands subject to actuation, following-distance, and field-of-view constraints. Through simulations and real-world experiments, we validate the effectiveness of the proposed framework and show it has better performance than existing frameworks.
\end{abstract}

\section{Introduction}

In applications such as ecological observation, inspection, and docking~\cite{tijjani2022,lwin2018}, underwater target tracking requires an AUV to maintain a desired relative position while keeping the moving target within its camera's field of view. Reliable close-range tracking, however, depends on consistently accurate estimates of the target's position and motion. While acoustic localization is well established and effective over medium and long ranges, it generally has limited spatial resolution and is susceptible to measurement noise and multipath interference at closer ranges\cite{huy2023}. These shortcomings are exacerbated when tracking highly maneuverable targets with unknown motion. Consequently, optical cameras are well suited to close-range underwater target tracking because they provide detailed target appearance and motion cues.

To translate these visual observations into vehicle commands, visual servoing is commonly categorized into image-based visual servoing (IBVS) and position-based visual servoing (PBVS) \cite{chaumette2006}. IBVS directly regulates image features, but changes in the shape, pose, and apparent size of deformable underwater targets weaken the relation between image-space error and physical distance. PBVS reconstructs the target's metric 3D relative state and regulates the vehicle in Cartesian space, allowing tracking distance and camera visibility to be handled explicitly. We therefore adopt PBVS for close-range underwater target tracking.

\begin{figure}[!t]
    \centering
    \includegraphics[width=\columnwidth]{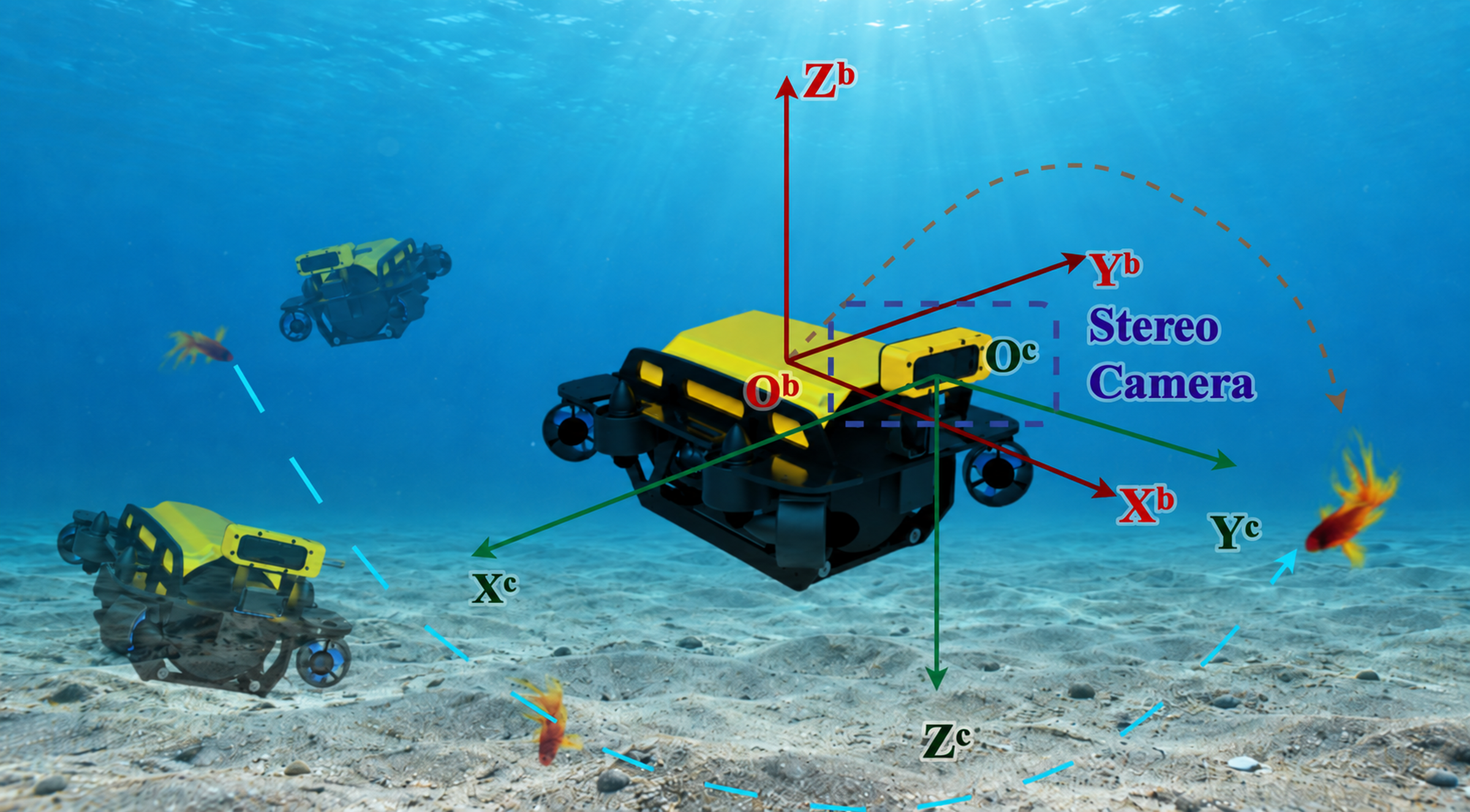}
    \vspace{-3mm}
    \caption{Schematic of vision-based underwater target tracking.}
    \label{fig:coordinate-frames}
    \vspace{-3mm}
\end{figure}


Reliable PBVS requires an accurate, target-specific metric depth estimate. Monocular cues such as bounding-box size provide only indirect range estimates and are sensitive to changes in target size, pose, and deformation. Stereo matching provides metric depth \cite{wu2019,lwin2018}, but converting a scene-level disparity map into a reliable target-depth measurement involves two challenges. First, underwater scattering, color attenuation, weak texture, and refraction can degrade stereo correspondence \cite{huy2023,akkaynak2018}. Second, stereo matching alone cannot determine which disparity measurements belong to the target. Detection boxes provide only coarse target regions and often include background pixels, while foreground--background disparity mixing near target boundaries can contaminate depth aggregation \cite{pon2020}. Background subtraction can refine the target region, but it may absorb hovering or slow-moving targets into the background model \cite{bouwmans2014}. We therefore introduce a target-depth mask that selects reliable target pixels before depth aggregation, thereby improving target-depth estimation and supporting robust 3D relative-state estimation.

Given the estimated relative state, the controller must track unknown target motion in real time while satisfying actuation, following-distance, and field-of-view constraints. Conventional PID, sliding-mode, and adaptive controllers generally do not optimize future tracking behavior under multiple constraints \cite{sha2025,meng2023}. Existing MPC approaches often focus on predefined trajectories or rely on a fixed target-motion model~\cite{zhang2019,shen2018,gao2024,gao2016}. These limitations, combined with the high computational cost of jointly optimizing yaw and translation, make it difficult to accommodate unknown target motion while meeting real-time control requirements. We therefore separate yaw regulation from translational control to enable efficient constrained translational MPC. The translational controller employs adaptive model-fusion predictive control, combining constant-velocity and zero-velocity target models with weights updated online from historical multi-step prediction errors. This online adaptation accommodates changes in target motion while reducing fixed-model lag and overshoot.

Our main contributions are summarized as follows:
\begin{enumerate}
\item We develop a target-specific perception method that combines color, disparity, and temporal cues to extract reliable target depth from scene-level stereo estimates. Separate filtering of image-plane and depth observations offers a stable estimate of the target's 3D relative state.

\item We design a real-time controller that decouples yaw regulation from translational control, enabling efficient QP-based translational MPC under multiple constraints. The controller adaptively combines constant-velocity and zero-velocity target models using weights updated from historical multi-step prediction errors, enabling adaptation to changing target motion.

\item We integrate the proposed estimator and controller into a complete PBVS underwater target-tracking system. Simulations and physical pool experiments on an eight-thruster AUV validate the perception module, controller, and overall closed-loop system.
\end{enumerate}

\section{Related Work}
\subsection{Underwater Target Perception}

Underwater target perception must provide reliable target localization and
metric depth despite degraded visibility and background interference
\cite{huy2023,alawode2022utb180,zhang2024webuot}. General-purpose
tracking-by-detection methods improve 2D target association but do not directly
provide target-specific metric depth \cite{zhang2022bytetrack}. Stereo-based
approaches combine target localization with geometric depth estimation. Wu et
al. combined YOLO with SGBM to detect fish and estimate their stereo depth
\cite{wu2019}, while Lwin et al. used dual-eye cameras to estimate the 3D pose
of a predefined docking marker \cite{lwin2018}. Object-centric stereo matching
and background modeling can reduce background interference
\cite{pon2020,bouwmans2014}. However, detector-defined regions may include
background pixels, while background models may fail for slowly moving targets.
Such errors in target-region selection can make depth aggregation unstable.

Accurate target segmentation alone does not guarantee reliable depth.
Underwater attenuation and backscatter can corrupt stereo correspondence,
producing invalid or noisy disparities even on target pixels
\cite{akkaynak2018}. UWStereo and Fast-FoundationStereo improve disparity
estimation through domain adaptation and efficient zero-shot inference
\cite{lv2025,fastfoundationstereo2026}, but their disparity maps may still contain
unreliable pixels in degraded underwater imagery. Consequently, segmentation
masks, including those generated by models such as SAM~2
\cite{ravi2025sam2}, cannot by themselves determine which target pixels have
disparities suitable for metric-depth estimation. This motivates our joint
use of color, disparity, and temporal cues to select reliable target pixels
before depth aggregation and filtering.

\begin{figure}[!t]
    \centering
    \includegraphics[width=\columnwidth]{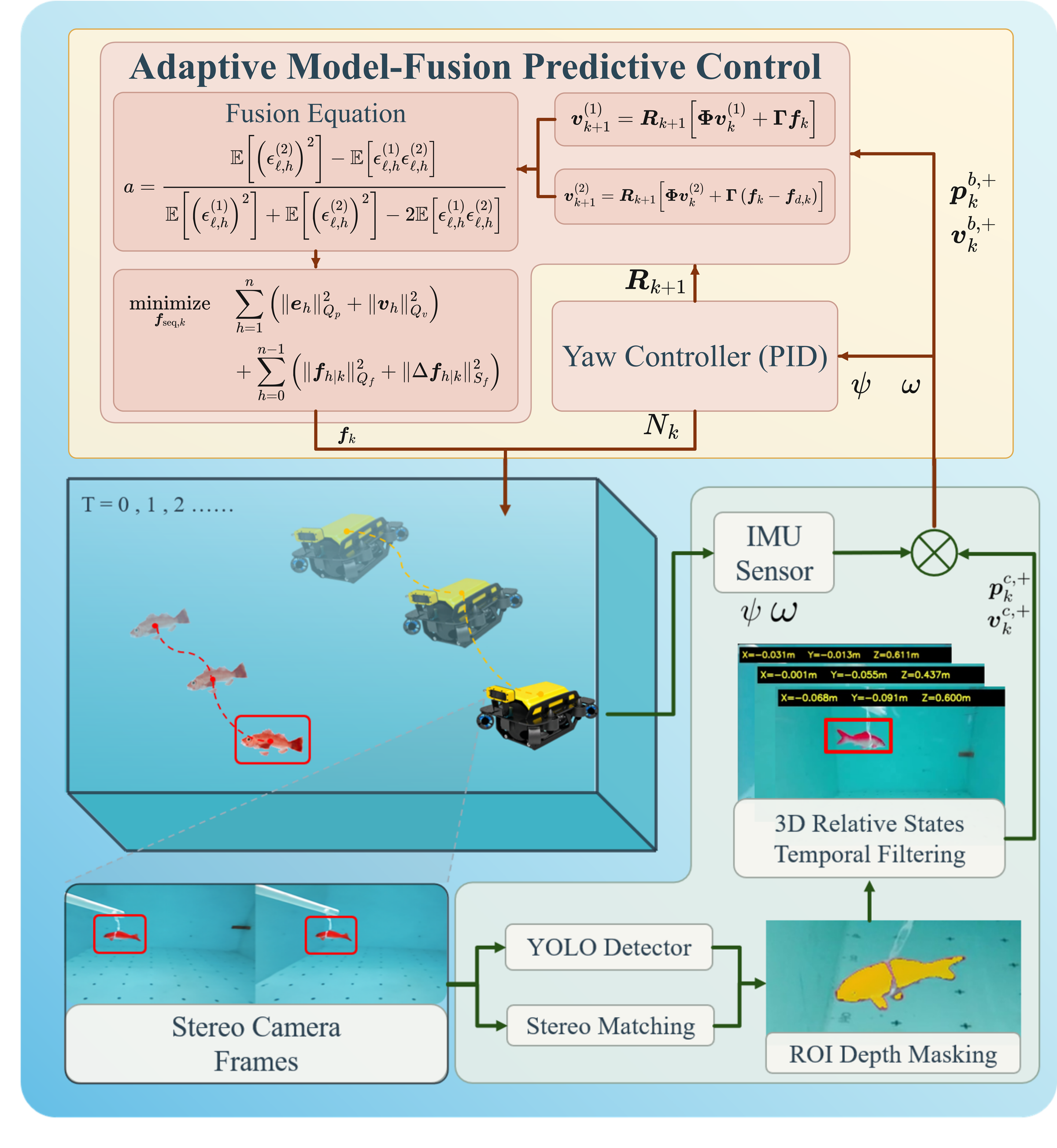}    \caption{Overall framework of the proposed underwater visual target-tracking system. Stereo images and IMU measurements provide a filtered 3D target-relative state; decoupled yaw control and adaptive model-fusion translational MPC generate constrained commands for the eight-thruster AUV.}
    \Description{Overall framework of stereo perception, state filtering,
    model-fusion MPC, and thruster control allocation.}
    \label{fig:framework}
\end{figure}

\subsection{Underwater Tracking Control}

Underwater target-following control must convert visual feedback into
vehicle motion while handling nonlinear dynamics, disturbances, and motion
coupling. Sha et al. developed a dual-loop controller for vision-based tracking with a multi-thruster AUV \cite{sha2025}. Meng et al. combined nonsingular terminal sliding-mode control with fuzzy inference to coordinate distance and azimuth tracking for a robotic manta \cite{meng2023}. These controllers demonstrate practical robustness but generate commands mainly from current tracking errors without explicitly predicting future target motion or constraint evolution.

Model predictive control (MPC) enables future motion to be optimized subject to system constraints. Zhang et al. formulated constrained 3D AUV trajectory tracking as a quadratic program \cite{zhang2019}, while Shen et al. developed a Lyapunov-based MPC scheme for stable trajectory tracking \cite{shen2018}. Gao et al. applied MPC to agile trajectory tracking of a fully vectored underwater robot \cite{gao2024}. For underwater visual servoing, Gao et al. combined nonlinear MPC for
constrained image-trajectory prediction with adaptive neural-network
compensation for uncertain vehicle dynamics but did not adapt the
target-motion model \cite{gao2016}; a related image-based MPC scheme
incorporated target-motion estimation under a single fixed motion model
\cite{liu2023ibvsmpc}. Adaptive multi-model estimation frameworks \cite{blom1988imm,mazor1998survey,na2022adaptive} mix motion
hypotheses using single-step likelihoods for state estimation; our
formulation instead weights models by historical multi-step prediction
errors and applies the fused prediction inside the control horizon.

\section{Methodology}

\subsection{Task Definition and System Framework}

This section formulates the target tracking task and defines the state interface between visual perception and control. For the task, the AUV is required to maintain a desired relative position with respect to the target while keeping the target within the camera field of view. The left-camera frame \(\{O_c-X_cY_cZ_c\}\), and the right-handed AUV body frame
\(\{O_b-X_bY_bZ_b\}\) are shown in Fig. \ref{fig:coordinate-frames}.

 For perception, the visual module estimates the target-relative position and velocity in the camera frame:
\[\boldsymbol{p}_k^{c,+}=\left[X_k^{c,+},Y_k^{c,+},Z_k^{c,+}\right]^{\mathrm{T}},\boldsymbol{v}_k^{c,+}=\left[\dot{X}_k^{c,+},\dot{Y}_k^{c,+},\dot{Z}_k^{c,+}\right]^{\mathrm{T}},\]
where the superscript \(+\) denotes the states after the decoupled image-center and depth filters.

For control, the state is first transformed into the body frame using the calibrated camera-to-body extrinsic parameters:
\begin{equation}
\widetilde{\boldsymbol p}_k^b
=
\boldsymbol r_{BC}^b+\boldsymbol R_c^b\boldsymbol p_k^{c,+}, \quad
\widetilde{\boldsymbol v}_k^b
=
\boldsymbol R_c^b\boldsymbol v_k^{c,+}.
\label{eq:transform}
\end{equation}
Here, \(\boldsymbol r_{BC}^b\) denotes the camera position relative to the body origin, and \(\boldsymbol R_c^b\) maps vectors from the camera frame to the body frame. The transformed state is then fused with the AUV attitude and angular-rate measurements to obtain the filtered body-frame state \(\boldsymbol p_k^{b,+}\) and \(\boldsymbol v_k^{b,+}\).

Let
$
\boldsymbol p^*
=
[p_x^*,p_y^*,p_z^*]^{\mathrm T}
$
denote the desired translational component of the relative pose in the body frame. The translational tracking error is defined as
\begin{equation}
\boldsymbol e_k^b
=
\boldsymbol p_k^{b,+}-\boldsymbol p^*.
\label{eq:error}
\end{equation}
The controller generates commands from the state defined in Eq.~\eqref{eq:error} while satisfying force, force-rate, following-distance, and field-of-view constraints. The complete architecture is summarized in Fig.~\ref{fig:framework}.

\subsection{Target-Specific State Estimation}

This section converts rectified stereo observations into a target-specific relative state for control. The pipeline reconstructs scene depth from stereo disparity, selects reliable target pixels using color, disparity, and temporal cues for target-specific depth measurement, and separately filters the depth and image center to estimate the target's 3D relative state.

\subsubsection{Stereo Geometric Measurement}

Let \(\mathcal I_k=(I_k^L,I_k^R)\) denote the rectified
left and right images at frame \(k\).Fast-FoundationStereo \cite{fastfoundationstereo2026} estimates the
left-referenced disparity map \(\mathbf D_k\in\mathbb R^{H\times W}\)
from rectified stereo images.
For a left-image pixel \(\boldsymbol\pi=(u,v)\), where \(u\) and \(v\)
are horizontal and vertical pixel coordinates, its disparity is
\(d_k(\boldsymbol\pi)=[\mathbf D_k]_{v,u}\).
Stereo triangulation and pinhole back-projection give
\cite{hartley2004}
\begin{equation}
\begin{gathered}
Z_k^c(\boldsymbol\pi)=\frac{f_xB}{d_k(\boldsymbol\pi)},\\
X_k^c(\boldsymbol\pi)=\frac{u-c_x}{f_x}Z_k^c(\boldsymbol\pi),\quad
Y_k^c(\boldsymbol\pi)=\frac{v-c_y}{f_y}Z_k^c(\boldsymbol\pi),
\end{gathered}
\label{eq:visual-stereo}
\end{equation}
where \(f_x,f_y\), \((c_x,c_y)\), and \(B\) are the calibrated focal lengths,
principal point, and the stereo baseline.

\begin{figure}[!t]
    \centering
    \includegraphics[width=\linewidth]{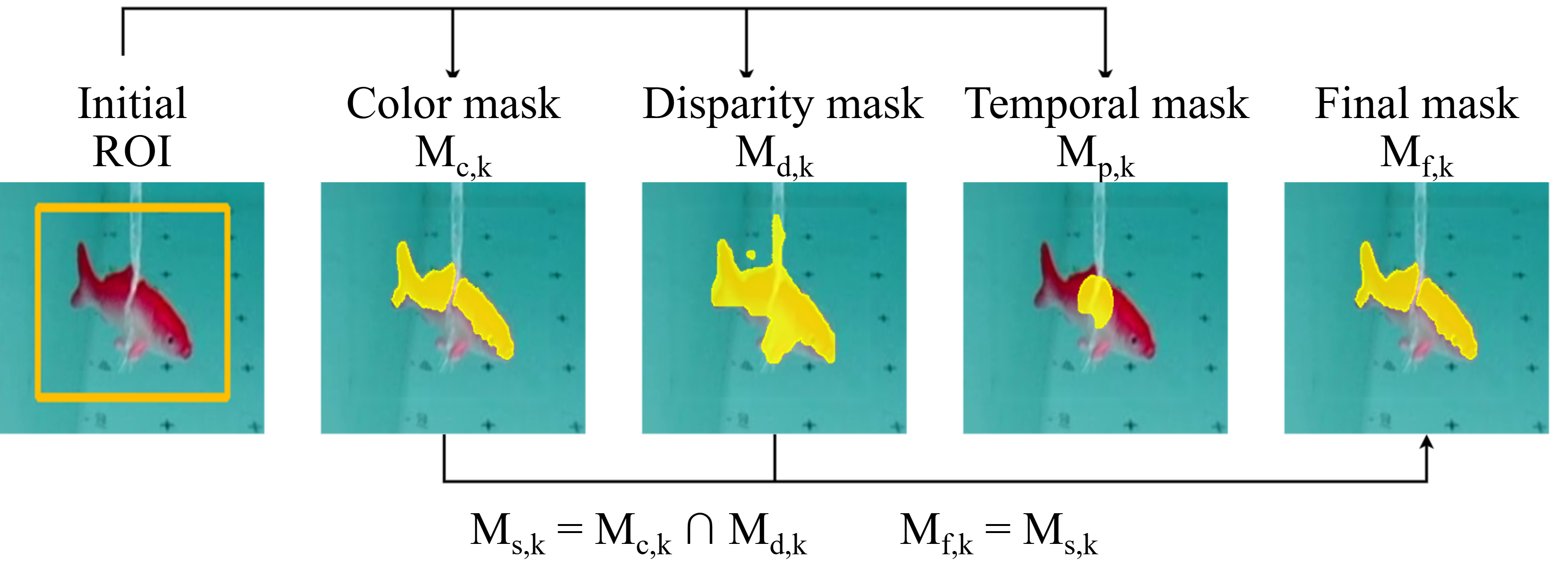}\\[0.6mm]
    {\footnotesize (a) Normal frame.}\\[1.2mm]

    \includegraphics[width=\linewidth]{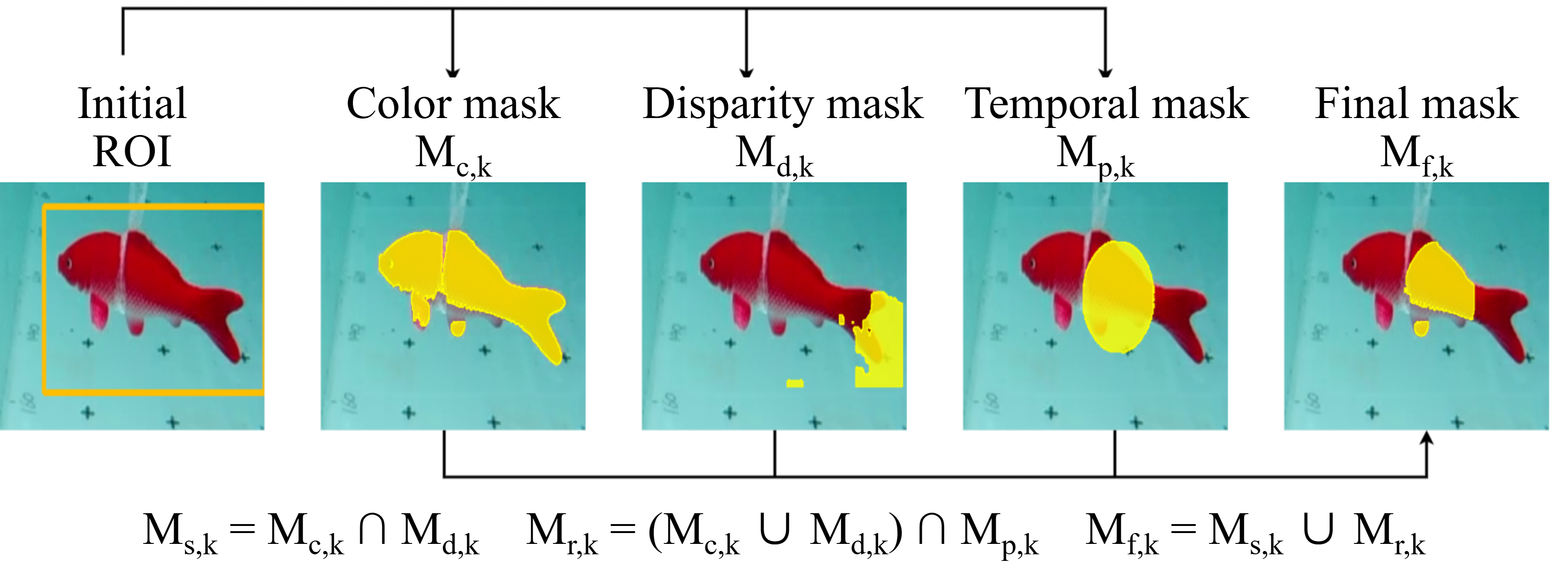}\\[0.6mm]
    {\footnotesize (b) Temporal recovery frame.}

    \caption{Target-depth mask generation in (a) a normal frame and (b) a
    temporal recovery frame with incomplete color--disparity overlap.}
    \Description{Target-depth mask generation for a normal valid frame and a temporal recovery frame.}
    \label{fig:mask-cases}
\end{figure}

\subsubsection{Target-Depth Mask}

Building on the 3D measurements obtained above, the target-depth mask combines color, disparity, and temporal-consistency cues. Color and disparity are first used to construct separate cue-specific masks, whose intersection forms a conservative joint mask. When the joint mask is too sparse, temporal support recovers additional pixels that remain consistent with the previous valid observation. Fig.~\ref{fig:mask-cases} illustrates both cases.

Let \(\mathcal R_k\) denote the target region and
\(\mathcal I_d\) the valid disparity interval. The valid pixels within the
target region are
\(\Omega_{v,k}=\{\boldsymbol\pi\in\mathcal R_k\mid
d_k(\boldsymbol\pi)\in\mathcal I_d\}\).
To obtain local background statistics, we expand the target region by a factor
\(s>1\) to form \(\mathcal R_k^{\mathrm{ext}}\), and define the surrounding
band as
\(\mathcal B_k=\mathcal R_k^{\mathrm{ext}}\setminus\mathcal R_k\).

We first construct a color-based mask. To avoid hue discontinuities, each pixel
is represented by its hue, saturation, and value as
\begin{equation}
\boldsymbol c_k(\boldsymbol{\pi})=
\begin{bmatrix}
S_k(\boldsymbol{\pi})\cos\theta_k(\boldsymbol{\pi})\\
S_k(\boldsymbol{\pi})\sin\theta_k(\boldsymbol{\pi})\\
V_k(\boldsymbol{\pi})
\end{bmatrix},
\quad
\theta_k(\boldsymbol{\pi})
=
\frac{2\pi H_k(\boldsymbol{\pi})}{180},
\end{equation}
where \(H_k\), \(S_k\), and \(V_k\) denote hue, saturation, and value, with
\(S_k,V_k\in[0,1]\). The mean
\(\boldsymbol\mu_k\) and covariance
\(\boldsymbol\Sigma_k\) of the background colors are estimated from
\(\mathcal B_k\). Covariance shrinkage improves the stability of the inverse
covariance \cite{ledoit2004}:
\begin{equation}
\widetilde{\boldsymbol\Sigma}_k
=
(1-\lambda)\boldsymbol\Sigma_k
+
\frac{\lambda}{3}
\operatorname{tr}(\boldsymbol\Sigma_k)I_3,
\end{equation}
where \(\lambda\in[0,1]\) controls the shrinkage strength. The color
abnormality score is
\begin{equation}
s_{c,k}(\boldsymbol{\pi})
=
(\boldsymbol c_k(\boldsymbol{\pi})-\boldsymbol\mu_k)^{\mathrm T}
\widetilde{\boldsymbol\Sigma}_k^{-1}
(\boldsymbol c_k(\boldsymbol{\pi})-\boldsymbol\mu_k).
\end{equation}
Otsu's method \cite{otsu1979} determines the color threshold \(\tau_{c,k}\),
then
\begin{equation}
\mathcal M_{c,k}
=
\left\{
\boldsymbol{\pi}\in\Omega_{v,k}
\mid
s_{c,k}(\boldsymbol{\pi})\geq\tau_{c,k}
\right\}.
\label{eq:visual-color-mask}
\end{equation}

We independently construct a disparity-based mask. Because the target is
generally closer than its local background, its pixels tend to have larger
disparities. Applying Otsu's method to the valid disparities gives the
threshold \(\tau_{d,k}\) and the mask
\begin{equation}
\mathcal M_{d,k}
=
\left\{
\boldsymbol{\pi}\in\Omega_{v,k}
\mid
d_k(\boldsymbol{\pi})\geq\tau_{d,k}
\right\}.
\label{eq:visual-disparity-mask}
\end{equation}
The two cue-specific masks are intersected to obtain the conservative joint
mask:
\begin{equation}
\mathcal M_{s,k}
=
\mathcal M_{c,k}\cap\mathcal M_{d,k}.
\label{eq:visual-joint-mask}
\end{equation}

When the joint mask is insufficient, temporal support uses the reference
from the most recent frame with an available depth measurement,
\(\boldsymbol q_{\mathrm{ref}}=[u_{\mathrm{ref}},v_{\mathrm{ref}},
d_{\mathrm{ref}}]^{\mathrm T}\), which contains the filtered image center and
median disparity. The normalized distance
between a current pixel and this reference is
\begin{equation}
\delta_k(\boldsymbol\pi)=
\frac{(u-u_{\mathrm{ref}})^2}{\sigma_u^2}
+\frac{(v-v_{\mathrm{ref}})^2}{\sigma_v^2}
+\frac{(d_k(\boldsymbol\pi)-d_{\mathrm{ref}})^2}{\sigma_d^2}.
\label{eq:visual-temporal-distance}
\end{equation}
Here, \(\sigma_u\), \(\sigma_v\), and \(\sigma_d\) normalize the horizontal,
vertical, and disparity deviations, and \(\gamma\) is the temporal-consistency
threshold. The temporal support is
\(\mathcal M_{p,k}=\{\boldsymbol\pi\in\Omega_{v,k}\mid
\delta_k(\boldsymbol\pi)\leq\gamma\}\), and the recovery region is
\(\mathcal M_{r,k}=(\mathcal M_{c,k}\cup\mathcal M_{d,k})
\cap\mathcal M_{p,k}\).

The final target-depth mask is
\begin{equation}
\mathcal M_{f,k}=
\begin{cases}
\mathcal M_{s,k},&|\mathcal M_{s,k}|\geq N_s,\\
\mathcal M_{s,k}\cup\mathcal M_{r,k},
&|\mathcal M_{s,k}|<N_s,\ |\mathcal M_{r,k}|\geq N_r,\\
\varnothing,
&|\mathcal M_{s,k}|<N_s,\ |\mathcal M_{r,k}|<N_r.
\end{cases}
\label{eq:visual-final-mask}
\end{equation}
Here, \(N_s\) and \(N_r\) are minimum pixel counts required for the joint
and recovery masks, respectively. Recovery therefore adds
only temporally consistent pixels retaining current-frame color or disparity
evidence. If
\(\mathcal M_{f,k}\) is nonempty, the valid depths within the final mask are
aggregated and will be filtered by the following method.

\subsubsection{Decoupled Filtering of Image-Center and Depth Measurements}

The detected image center and target depth are filtered independently before
3D state reconstruction. For the depth measurement, a constant-velocity
Kalman filter \cite{kalman1960} uses the state
\(\boldsymbol\zeta_k=
[Z_k^c,\dot Z_k^c]^{\mathsf T}.\)
Its prediction model is
\begin{equation}
\boldsymbol\zeta_k^-
=
F_k\boldsymbol\zeta_{k-1}^+,
\quad
F_k=
\begin{bmatrix}
1&\Delta t_k\\
0&1
\end{bmatrix}.
\label{eq:visual-depth-prediction}
\end{equation}
The measurement model is
\begin{equation}
z_k
=
H_Z\boldsymbol\zeta_k+\varepsilon_{z,k},
\quad
H_Z=[1,0],
\varepsilon_{z,k}\sim\mathcal N(0,R_{z,k}).
\label{eq:visual-depth-observation}
\end{equation}
The process covariance combines a random-acceleration model
\cite{barshalom2002} with depth diffusion:
\begin{equation}
Q_k
=
\sigma_a^2
\begin{bmatrix}
\Delta t_k^4/4&\Delta t_k^3/2\\
\Delta t_k^3/2&\Delta t_k^2
\end{bmatrix}
+
\begin{bmatrix}
q_z\max(\Delta t_k,\Delta t_0)&0\\
0&0
\end{bmatrix}.
\label{eq:visual-depth-process-noise}
\end{equation}

Let \(n_k=|\mathcal M_{f,k}|\). The robust scale of the selected depth samples
and the adaptive measurement variance are \cite{huber2009}
\begin{equation}
\widehat\sigma_{Z,k}=0.7413\,\operatorname{IQR}
\left(\{Z_k^c(\boldsymbol\pi)\}_{\boldsymbol\pi\in\mathcal M_{f,k}}\right),
R_{z,k}=\eta_R\frac{\pi}{2n_k}\widehat\sigma_{Z,k}^2.
\label{eq:visual-depth-variance}
\end{equation}
The factor \(0.7413\) converts IQR to a robust standard-deviation estimate,
and \(\eta_R\) is a tuning coefficient. More valid depth samples and lower depth dispersion correspond
to a smaller measurement variance \(R_{z,k}\). The Kalman update
is applied when a valid depth measurement is available; otherwise,
\(\boldsymbol\zeta_k^+=\boldsymbol\zeta_k^-\).

For the detected center \(\boldsymbol\xi_k=[u_k,v_k]^{\mathrm T}\), an
\(\alpha\)-\(\beta\) filter outputs
\(\boldsymbol\xi_k^+=[u_k^+,v_k^+]^{\mathrm T}\) and
\(\dot{\boldsymbol\xi}_k^+=[\dot u_k^+,\dot v_k^+]^{\mathrm T}\), and continues
to update when depth is unavailable.

Let \(\boldsymbol s_k^+=[u_k^+,v_k^+,Z_k^{c,+}]^{\mathrm T}\) and
\(\dot{\boldsymbol s}_k^+=[\dot u_k^+,\dot v_k^+,\dot Z_k^{c,+}]^{\mathrm T}\).
Define the pinhole mapping
\(g(u,v,Z)=[(u-c_x)Z/f_x,(v-c_y)Z/f_y,Z]^{\mathrm T}\).
The reconstructed state satisfies:
\begin{equation}
\begin{gathered}
\boldsymbol p_k^{c,+}=g(\boldsymbol s_k^+),\quad
\boldsymbol v_k^{c,+}=J_g(\boldsymbol s_k^+)\dot{\boldsymbol s}_k^+,\\
J_g(u,v,Z)=
\begin{bmatrix}
Z/f_x&0&(u-c_x)/f_x\\
0&Z/f_y&(v-c_y)/f_y\\
0&0&1
\end{bmatrix}.
\end{gathered}
\label{eq:visual-state-reconstruction}
\end{equation}
The filtered
camera-frame state is then transformed into the body frame based on Eq. \eqref{eq:transform} and provided to the
controller.

In addition, the filtered image center and the median disparity \(\bar d_k=\operatorname{median}_{\boldsymbol\pi\in\mathcal M_{f,k}}
d_k(\boldsymbol\pi)\) will update the reference \(\boldsymbol q_{\mathrm{ref}}=[u_k^+,v_k^+,\bar d_k]^{\mathrm T}\)
used for temporal mask recovery when a valid depth measurement is available.
Otherwise, the previous reference is retained. The overall state estimation procedure is organized in Algorithm \ref{alg:visual}.
\begin{algorithm}[t]
\caption{Online Stereo Relative-State Estimation}
\label{alg:visual}
\begin{algorithmic}[1]
\STATE \textbf{Require:}
\(\mathcal I_k,\mathcal R_k,\boldsymbol\xi_k,\Delta t_k\).
\STATE \textbf{Ensure:}
\(\boldsymbol p_k^{c,+},\boldsymbol v_k^{c,+}\).

\STATE \(\mathbf D_k\gets
\textsc{FastFoundationStereo}(\mathcal I_k)\)
\cite{fastfoundationstereo2026}.
\STATE \((\Omega_{v,k},\mathcal B_k)\gets
\textsc{ValidROI}(\mathbf D_k,\mathcal R_k)\).
\STATE \((\mathcal M_{c,k},\mathcal M_{d,k})\gets
\textsc{Masks}\) using
\eqref{eq:visual-color-mask}--\eqref{eq:visual-disparity-mask}.
\STATE \(\mathcal M_{s,k}\gets\textsc{JointMask}\)
using \eqref{eq:visual-joint-mask}.
\STATE \(\mathcal M_{f,k}\gets\textsc{RecoverMask}\)
using \eqref{eq:visual-temporal-distance}--\eqref{eq:visual-final-mask}.

\IF{\(\textsc{DepthAvailable}(\mathcal M_{f,k},\mathbf D_k)\)}
    \STATE \(z_k\gets
    \operatorname{median}_{\boldsymbol\pi\in\mathcal M_{f,k}}
    Z_k^c(\boldsymbol\pi)\) using \eqref{eq:visual-stereo}.
    \STATE \(\bar d_k\gets
    \operatorname{median}_{\boldsymbol\pi\in\mathcal M_{f,k}}
    d_k(\boldsymbol\pi)\).
\ELSE
    \STATE \(z_k\gets\varnothing,\quad\bar d_k\gets\varnothing\).
\ENDIF

\STATE \((\boldsymbol\xi_k^+,\dot{\boldsymbol\xi}_k^+)
\gets\alpha\textnormal{-}\beta(\boldsymbol\xi_k)\).
\STATE \(\boldsymbol\zeta_k^+\gets\textsc{KalmanUpdate}\)
using \eqref{eq:visual-depth-prediction}--\eqref{eq:visual-depth-variance}.
\STATE \((\boldsymbol p_k^{c,+},\boldsymbol v_k^{c,+})
\gets\textsc{Reconstruct}\)
using \eqref{eq:visual-state-reconstruction}.
\IF{\(\bar d_k\neq\varnothing\)}
    \STATE \(\boldsymbol q_{\mathrm{ref}}\gets
    [u_k^+,v_k^+,\bar d_k]^{\mathrm T}\).
\ENDIF
\STATE \textbf{return}
\(\boldsymbol p_k^{c,+},\boldsymbol v_k^{c,+}\).
\end{algorithmic}
\end{algorithm}

\subsection{Controller Design}

At each visual update \(k\), the controller receives the tracking error based on Eq. \eqref{eq:error} through the above perception module. The controller separates yaw and translational control. A HOLD/TURN state machine manages yaw, while the
translational MPC predicts relative motion with two models by fusing their predictions using the online weight, and computes constrained translational commands. All variables in this section are expressed in the body frame, and the
superscript \(b\) is omitted. The current control-side filtered position and
velocity are denoted as \(\boldsymbol p_k^+\) and \(\boldsymbol v_k^+\).

\subsubsection{Yaw Control and Rotation Compensation}
To coordinate with the MPC, the yaw controller augments the cascaded PID with a state machine and rotation-matrix prediction, allowing it to adjust the yaw angle for target tracking while maintaining decoupled rotational and translational control.

Define the horizontal line-of-sight angle
\(\alpha_k=\operatorname{atan2}(p_{y,k}^{+},p_{x,k}^{+})\).
A hysteretic HOLD/TURN logic enters TURN when
\(|\alpha_k|>\alpha_{\mathrm{on}}\) for a preset dwell time and sets
\(\psi_k^{*}=\psi_k+\alpha_k\); it returns to HOLD when
\(|\alpha_k|<\alpha_{\mathrm{off}}\), where
\(\alpha_{\mathrm{off}}<\alpha_{\mathrm{on}}\), and freezes the yaw reference.
An IMU-based cascaded PID independently tracks \(\psi_k^{*}\) and generates the yaw torque \(\hat N_k\).

Because the relative state is expressed in the body frame, the yaw-induced frame rotation must be included in the translational prediction. Using the first-order yaw dynamics
\(m_w\dot{\omega}+d_w\omega=N\) \cite{fossen2011}, and holding \(\hat N_k\) and the state-machine mode fixed during each QP solve, we obtain:
\begin{equation}
\hat{\omega}_{h+1|k}
=
a_\omega\hat{\omega}_{h|k}
+
b_\omega\hat N_{h|k},
\end{equation}
where
\begin{equation}
a_\omega=
\exp\!\left(-\frac{d_wT_s}{m_w}\right),
\qquad
b_\omega=
\frac{1-a_\omega}{d_w}.
\end{equation}
\(\hat{\omega}_{h|k}\) is the predicted angular velocity at time \(k+h\), calculated at time \(k\). The corresponding yaw increment and passive body-frame rotation are
\begin{equation}
\begin{aligned}
\Delta\hat{\psi}_{h|k}
&=
\frac{T_s}{2}
\left(
\hat{\omega}_{h|k}+\hat{\omega}_{h+1|k}
\right),\\
\hat R_{h|k}
&=
R_z\!\left(-\Delta\hat{\psi}_{h|k}\right),
\qquad h=0,\ldots,n-1 .
\end{aligned}
\label{eq:yaw-rot}
\end{equation}
Here, \(m_w\), \(d_w\), \(T_s\), and \(n\) denote the effective inertia, damping, sampling interval, and prediction-horizon length, respectively. The rotation sequence is computed before the translational QP and held fixed during the solve, preserving linearity and convexity.

\subsubsection{Adaptive Model-Fusion Predictive Control}
The controller uses two models to predict relative motion and combines their predictions through a weighted sum. The weights are calculated from the accuracy of each model's past predictions. The resulting fused prediction is then used in the MPC cost function to compute the control force.

At the current time \(k\), Model 1 assumes a stationary target, while Model 2 assumes a moving target. Under Model 2, more thrust is needed to reduce the distance to the target. This additional force is denoted by \(\boldsymbol f_{d,k}\). The relative-velocity and position updates are given below:
\begin{align}
\boldsymbol v_{k+1}^{(1)}
&=
\boldsymbol R_{k+1}
\Bigl[
  \boldsymbol{\Phi}\boldsymbol v_k^{(1)}
  +
  \boldsymbol{\Gamma}\boldsymbol f_k
\Bigr],
\label{eq:model-1}
\\[3pt]
\boldsymbol v_{k+1}^{(2)}
&=
\boldsymbol R_{k+1}
\Bigl[
  \boldsymbol{\Phi}\boldsymbol v_k^{(2)}
  +
  \boldsymbol{\Gamma}
  \bigl(\boldsymbol f_k-\boldsymbol f_{d,k}\bigr)
\Bigr],
\label{eq:model-2}
\\[3pt]
\boldsymbol p_{k+1}^{(m)}
&=
\boldsymbol R_{k+1}\boldsymbol p_k^{(m)}
+
T_s\boldsymbol v_k^{(m)},
\qquad m\in\{1,2\}.
\end{align}
Here, \(\boldsymbol f_k\) is the driving force applied to AUV.\(\boldsymbol f_{d,k}\) is updated by smoothing \(\boldsymbol f_{k-1}\).

The translational dynamics follow a low-speed model with linearized damping, \(\boldsymbol{M_t}\dot{\boldsymbol v}+\boldsymbol{D_L}\boldsymbol v=\boldsymbol f\) \cite{fossen2011}. Using a first-order Taylor approximation, we obtain: 
\begin{equation}
\boldsymbol\Phi=e^{\boldsymbol\Lambda T_s},
\qquad
\boldsymbol\Gamma
=
-\int_0^{T_s}
e^{\boldsymbol\Lambda s}\boldsymbol M_t^{-1}\,\mathrm{d}s,
\end{equation}
where \(\boldsymbol\Lambda=-\boldsymbol M_t^{-1}\boldsymbol D_L\),
\(\boldsymbol M_t\) is the translational inertia including added mass,
\(\boldsymbol D_L\) is the linear damping matrix, and \(T_s\) is the sampling period. 
\begin{figure}[!t]
    \centering
    \includegraphics[width=\columnwidth]{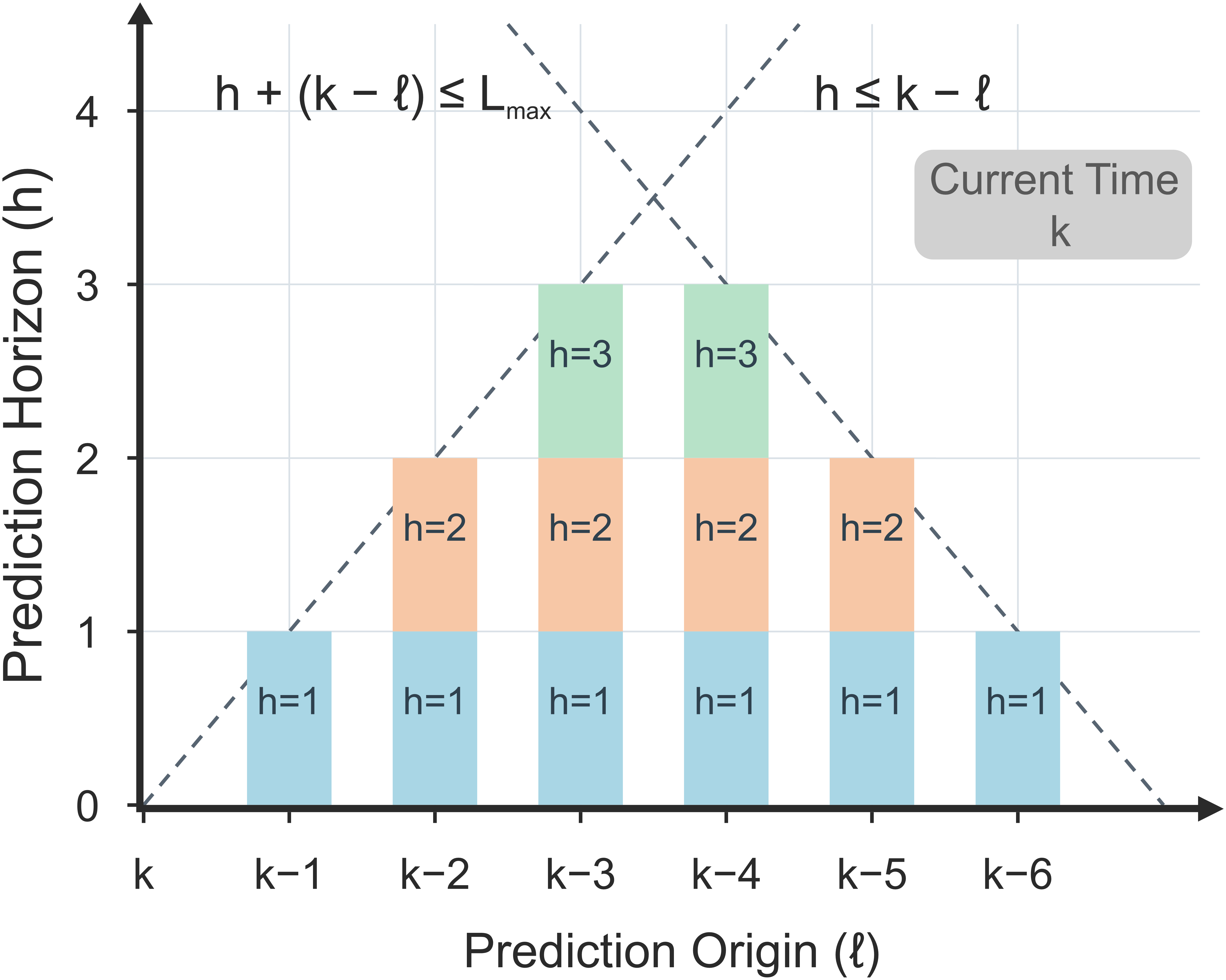}
 \caption{Sampling window for the fusion-weight update. Selected data pairs must satisfy two conditions. The condition $h \leq k-\ell$ ensures that the prediction--observation pair is available. The condition $h+(k-\ell) \leq L_{\max}$ excludes pairs with a large combined prediction age and horizon.($L_{\max}=7$ in this illustration).}
\Description{Schematic of the admissible sampling window for fusion-weight
  estimation: prediction-origin distance on the horizontal axis, prediction
  horizon on the vertical axis, cells marking usable completed
  prediction--observation pairs, and two dashed boundary constraints.}
    \label{fig:history_prediction}
\end{figure}

The historical prediction error of model \(m\) is defined as
\begin{equation}
\boldsymbol{\epsilon}_{\ell,h}^{(m)}
=
\boldsymbol p_{\ell+h}^{+}
-
\hat{\boldsymbol p}_{\ell+h\mid\ell}^{(m),\mathrm{eva}},
\qquad m\in\{1,2\}.
\label{eq:historical-model-error}
\end{equation}
It is used to evaluate the accuracy of the model. Here,
\(\hat{\boldsymbol p}_{\ell+h\mid\ell}^{(m),\mathrm{eva}}\)
is the \(h\)-step relative-position prediction starting from time \(\ell\) of model \(m\). 
 The controller selects several sets of past data for this evaluation.
Each set is specified by a prediction start time \(\ell\) and a prediction
horizon \(h\).

For each set, model \(m\) starts from the state at time \(\ell\). The model reads the driving force applied to the AUV at each time step from \(\ell\) to \(\ell+h\)  (\(\boldsymbol f_\ell,\ldots,\boldsymbol f_{\ell+h}\)). 
Based on these forces, the model predicts the
target's relative position at time \(\ell+h\). This prediction is compared
with the measured relative position \(\boldsymbol p_{\ell+h}^{+}\) . The pairs of \(\ell\) and \(h\) used for validation are selected
as shown in Fig.~\ref{fig:history_prediction}. 

The fused error \(e_{\ell,h}(a)\) is defined as
\begin{equation}
e_{\ell,h}(a)
=
a\epsilon_{\ell,h}^{(1)}
+
(1-a)\epsilon_{\ell,h}^{(2)}.
\end{equation}
Here, \(\epsilon_{\ell,h}^{(m)}\) is one scalar component of
\(\boldsymbol{\epsilon}_{\ell,h}^{(m)}\), and \(a\) is its fusion weight. The expected squared error is
\begin{equation}
\begin{aligned}
\mathbb{E}\!\Bigl[e_{\ell,h}^{2}(a)\Bigr]
={}&
a^{2}
\mathbb{E}\!\Bigl[
  \bigl(\epsilon_{\ell,h}^{(1)}\bigr)^{2}
\Bigr]
+
(1-a)^{2}
\mathbb{E}\!\Bigl[
  \bigl(\epsilon_{\ell,h}^{(2)}\bigr)^{2}
\Bigr]
\\
&+
2a(1-a)
\mathbb{E}\!\Bigl[
  \epsilon_{\ell,h}^{(1)}
  \epsilon_{\ell,h}^{(2)}
\Bigr].
\end{aligned}
\label{eq:fused-error-moment}
\end{equation}
Minimizing \(\mathbb{E}\!\bigl[e_{\ell,h}^2(a)\bigr]\) gives
\begin{equation}
a
=
\frac{
\mathbb{E}\!\Bigl[
\bigl(\epsilon_{\ell,h}^{(2)}\bigr)^2
\Bigr]
-
\mathbb{E}\!\Bigl[
\epsilon_{\ell,h}^{(1)}
\epsilon_{\ell,h}^{(2)}
\Bigr]
}{
\mathbb{E}\!\Bigl[
\bigl(\epsilon_{\ell,h}^{(1)}\bigr)^2
\Bigr]
+
\mathbb{E}\!\Bigl[
\bigl(\epsilon_{\ell,h}^{(2)}\bigr)^2
\Bigr]
-
2\mathbb{E}\!\Bigl[
\epsilon_{\ell,h}^{(1)}
\epsilon_{\ell,h}^{(2)}
\Bigr]
}.
\label{eq:weight}
\end{equation}
The fused predictions use the diagonal weight matrix \(\boldsymbol a\):\begin{equation}
\begin{aligned}
\hat{\boldsymbol p}
&=
\boldsymbol a\hat{\boldsymbol p}^{(1)}
+
(\boldsymbol I-\boldsymbol a)\hat{\boldsymbol p}^{(2)},\\
\hat{\boldsymbol v}
&=
\boldsymbol a\hat{\boldsymbol v}^{(1)}
+
(\boldsymbol I-\boldsymbol a)\hat{\boldsymbol v}^{(2)}.
\end{aligned}
\end{equation}
At time \(k\), the MPC solves the following quadratic program:
\begin{equation}
\begin{aligned}
\underset{\boldsymbol f_{\mathrm{seq},k}}{\operatorname{minimize}}
\quad &
\phantom{+}\sum_{h=1}^{n}
\left(
\lVert\boldsymbol e_h\rVert_{Q_p}^{2}
+
\lVert\boldsymbol v_h\rVert_{Q_v}^{2}
\right)
\\[-0.2em]
&+
\sum_{h=0}^{n-1}
\left(
\lVert\boldsymbol f_{h|k}\rVert_{Q_f}^{2}
+
\lVert\Delta\boldsymbol f_{h|k}\rVert_{S_f}^{2}
\right)
\\[0.3em]
\text{subject to}\quad &
\boldsymbol f_{\min}
\leq \boldsymbol f_{h|k}
\leq \boldsymbol f_{\max},
\\
&
\Delta\boldsymbol f_{\min}
\leq \Delta\boldsymbol f_{h|k}
\leq \Delta\boldsymbol f_{\max},
\\[-0.2em]
&
h=0,\ldots,n-1 .
\end{aligned}
\label{eq:cost}
\end{equation}
Here,
\(\boldsymbol f_{\mathrm{seq},k}
=
\operatorname{col}\{
\boldsymbol f_{0|k},\ldots,
\boldsymbol f_{n-1|k}\}\)
is the stacked force sequence, and
\(\lVert\boldsymbol x\rVert_{W}^{2}
\triangleq
\boldsymbol x^{\mathsf T}W\boldsymbol x\).
Moreover,
\(\boldsymbol e_h=\hat{\boldsymbol p}_{h|k}-\boldsymbol p^{*}\),
\(\Delta\boldsymbol f_{h+1|k}
=
\boldsymbol f_{h+1|k}-\boldsymbol f_{h|k}\).
The matrices \(Q_p\), \(Q_v\), \(Q_f\), and \(S_f\) are the
corresponding penalty weights, and all inequalities are elementwise.
The optimal force is combined with the attitude torques and allocated
to the eight thrusters.

\section{Experiments and Results}
\subsection{Platform and Experimental Setup}
We conducted a series of video-replay, simulation, and real
tracking experiments using an eight-thruster AUV \cite{xu2025aucamp} in a
\(4\,\mathrm m\times2\,\mathrm m\times1\,\mathrm m\) pool, assessing
target-depth measurement quality, relative-state stability, and
tracking accuracy.

The AUV carries a forward-looking stereo camera, and a fish-shaped model serves
as the tracking target. Four horizontal thrusters
control planar motion and yaw, while four vertical thrusters
control heave, roll, and pitch. An external host computer runs perception, state estimation, and
upper-level control through ROS~2, while the onboard controller handles attitude
control and thruster actuation. Controller simulations are performed in a
Unity-based underwater environment integrated with ROS~2. All methods use the same inputs and initial conditions, with calibrated camera parameters and fixed algorithm
parameters.

\subsection{Target-Depth Measurement and State-Filtering Evaluation}

Two replay experiments are conducted on the same 12 real-world videos, comprising four
static-disturbance and eight dynamic sequences. The dynamic sequences cover
target and vehicle motion, attitude changes, and target yaw. All methods share
the same rectified images, detection boxes, and disparity maps, while the same
final validity criteria are applied to every method. Each metric is first
computed per video and then summarized by its median across the 12 videos.

\subsubsection{Mask comparison}
This experiment evaluates target-depth masking without output
filtering; output filtering is disabled for all methods. A frame is considered
valid if the selected mask covers at
least \(5\%\) of the detection box, provides a finite median depth within
\(0.1\)--\(5\) m, and has a depth interquartile range (Z-IQR) no greater
than \(0.20\) m. The valid-frame rate (VR) is the fraction of
valid frames in each video. We compare the complete mask
\(\mathcal M_f\) (Ours-T), its single-frame variant without temporal
recovery (Ours-S), and the disparity-only ablation (Ours-D), together
with BBox, GrabCut \cite{rother2004grabcut}, and MOG2
\cite{zivkovic2004mog}. Table~\ref{tab:mask} summarizes the quantitative
results.

\begin{figure*}[!t]
    \centering
    \includegraphics[trim=0 126 0 95, clip, width=\textwidth]{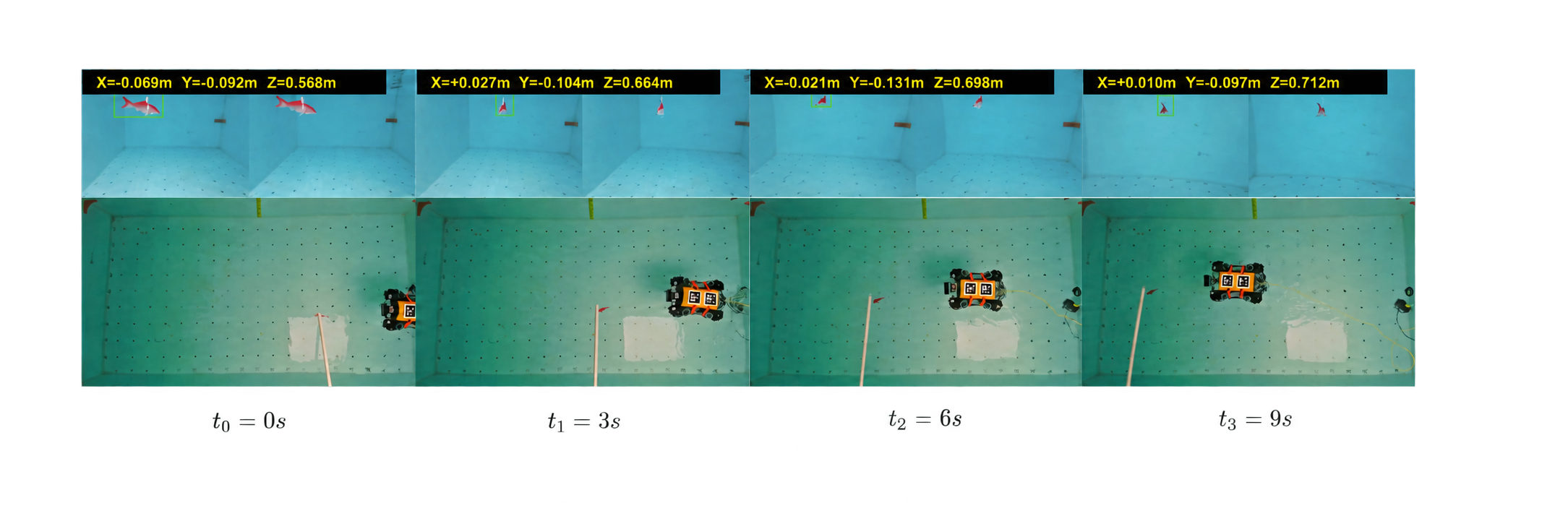}
    \caption{Closed-loop pool tracking at $t=0$, $3$, $6$, and $9\,\mathrm{s}$:
    onboard detections and filtered relative positions (top), with corresponding
    overhead views (bottom).}
    \Description{Four time-ordered onboard and overhead image pairs showing
    the AUV following a fish-shaped target in the pool.}
    \label{fig:tracking-motion}
\end{figure*}

\begin{figure}[!tbp]
    \centering
    \includegraphics[width=\linewidth]{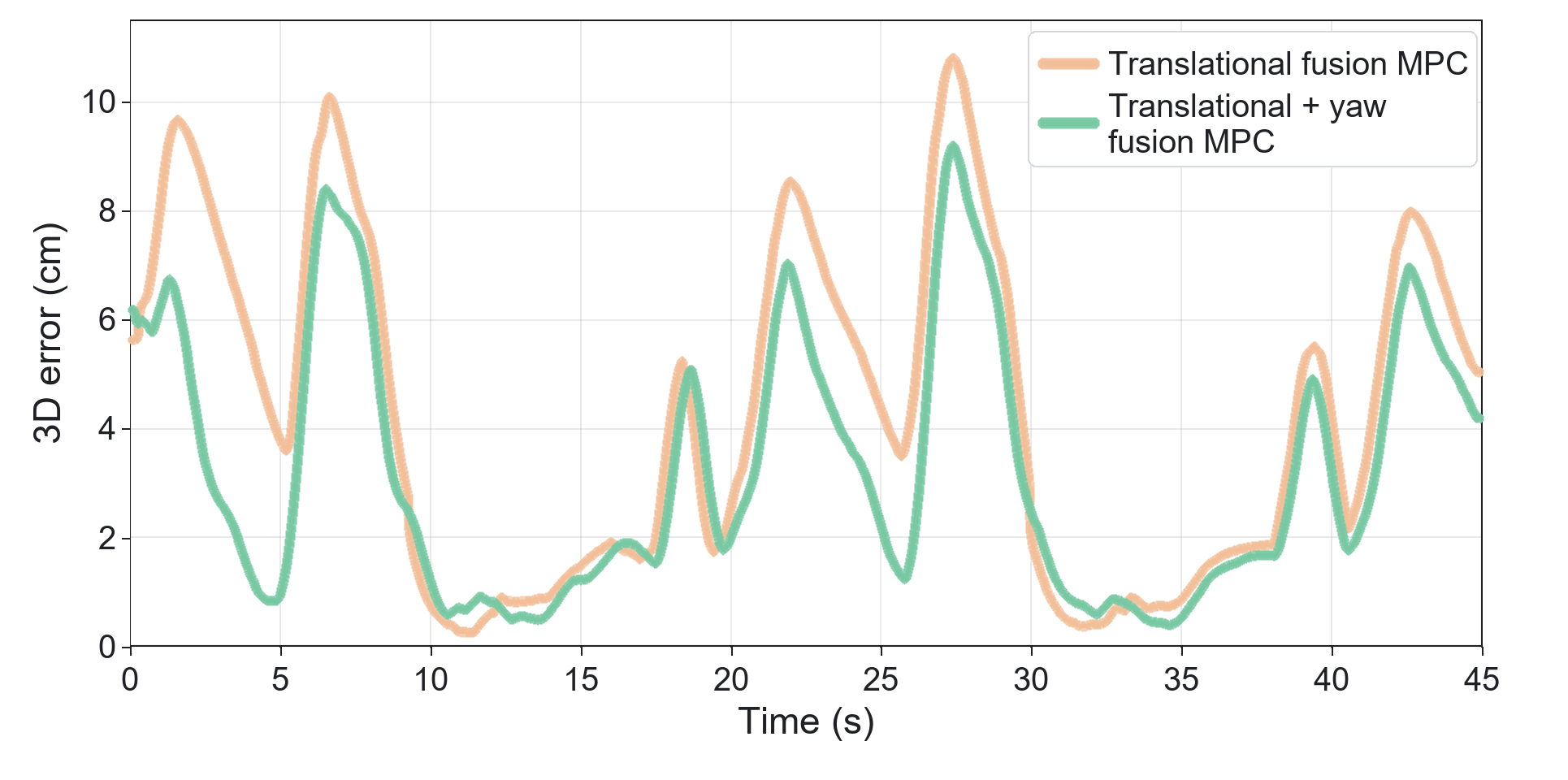}
    \caption{Position-error norms with and without yaw control in the
    rectangular-motion simulation.}
    \Description{Position-error norm over time for the model-fusion MPC with and
    without yaw control in the rectangular-motion Unity simulation.}
    \label{fig:rectangular-error}
\end{figure}

Ours-T achieves the highest median per-video VR of \(98.40\%\) and the lowest
macro-median Z-IQR of \(0.0150\,\mathrm m\). In contrast, the conventional
baselines remain below \(31\%\) VR and have macro-median Z-IQR values of at
least \(0.2435\,\mathrm m\). GrabCut requires \(178.50\,\mathrm{ms}\) per frame,
about \(30\times\) the time required by Ours-T.

\subsubsection{Mask and decoupled image-center/depth filtering}
\begin{figure}[!t]
    \centering
    \includegraphics[width=1\linewidth]{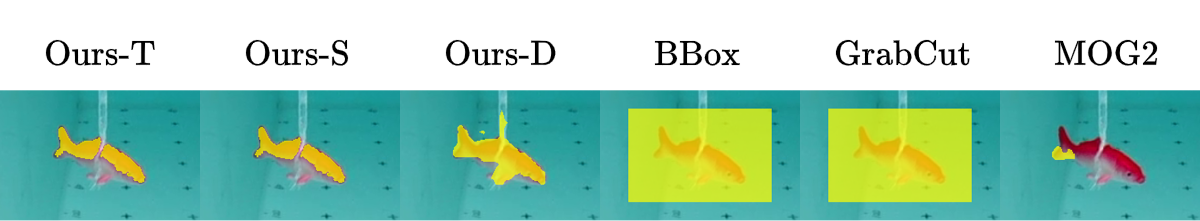}
    \caption{Target-depth mask on a representative fish sample.
    The proposed method suppresses background and mismatched disparities,
    while temporal support recovers short missing regions.}
    \Description{Target-depth masking result on a representative fish sample.}
    \label{fig:fish-mask-examples}
\end{figure}
This experiment separates the contributions of the two stages that stabilize the
relative state. BBox + Raw aggregates all valid disparities inside the detection
box, \(\mathcal M_f\) + Raw restricts the aggregation to the proposed
target-depth mask, and \(\mathcal M_f\) + Filters additionally applies the
depth Kalman filter and image-center \(\alpha\)-\(\beta\) filter.
The metrics are the P95 frame-to-frame changes of the
target depth, of the image-plane center coordinates, and of the line-of-sight
(LOS) angle. Both raw configurations use the same unfiltered detection-box
center for the image-plane measurements; therefore, the mask affects only the
depth channel in this comparison.

\begin{table}[!t]
\caption{Macro-medians of per-video P95 frame-to-frame changes in the replay experiments.}
\label{tab:mask-stages}
\centering
\small
\setlength{\tabcolsep}{4pt}
\begin{tabular*}{\columnwidth}{@{\extracolsep{\fill}}lcccc@{}}
\toprule
\textbf{Configuration} & \(\Delta Z/\mathrm{m}\,\downarrow\) &
\(\Delta u/\mathrm{px}\,\downarrow\) &
\(\Delta v/\mathrm{px}\,\downarrow\) &
\(\mathrm{LOS}/{}^{\circ}\,\downarrow\)\\
\midrule
BBox + Raw & 0.7161 & 5.350 & 4.250 & 0.5667\\
\(\mathcal M_f\) + Raw & 0.0449 & 5.350 & 4.250 & 0.5667\\
\(\mathcal M_f\) + Filters & \textbf{0.0264} & \textbf{5.175} & \textbf{3.293} & \textbf{0.4896}\\
\bottomrule
\end{tabular*}
\end{table}

\begin{table}[!t]
\caption{Macro-median target-depth results. VR denotes the valid-frame
rate, Z-IQR the depth interquartile range, \(\Delta Z\) the frame-to-frame
depth change, and P95 the 95th percentile.}
\label{tab:mask}
\centering
\scriptsize
\setlength{\tabcolsep}{1.5pt}
\renewcommand{\arraystretch}{1.10}
\begin{tabular*}{\columnwidth}{@{\extracolsep{\fill}}lccccc@{}}
\toprule
\textbf{Method} &
\textbf{VR} \(\uparrow\) &
\textbf{IQR-Med.} (m) \(\downarrow\) &
\textbf{IQR-P95} (m) \(\downarrow\) &
\(\boldsymbol{\Delta Z}\)-\textbf{P95} (m) \(\downarrow\) &
\textbf{Time} (ms) \(\downarrow\) \\
\midrule
Ours-T & \textbf{0.9840} & \textbf{0.0150} &
\textbf{0.0592} & \textbf{0.0449} & 5.92 \\
Ours-S & 0.9819 & 0.0152 & 0.0593 & 0.0452 & 5.15 \\
Ours-D & 0.9821 & 0.0304 & 0.1325 & 0.0879 & \textbf{0.24} \\
\midrule
BBox & 0.2891 & 0.2435 & 0.9383 & 0.7161 & 0.41 \\
GrabCut & 0.2876 & 0.2443 & 1.0025 & 0.7099 & 178.50 \\
MOG2 & 0.3073 & 0.3616 & 0.9079 & 0.1121 & 7.95 \\
\bottomrule
\end{tabular*}
\end{table}

Table~\ref{tab:mask-stages} shows that the mask provides most of the depth
stabilization: replacing the box aggregate by \(\mathcal M_f\) cuts the
macro-median P95 \(\Delta Z\) from \(0.7161\,\mathrm m\) to
\(0.0449\,\mathrm m\), a factor of about \(16\), because the background
disparities that dominate the box are removed before the median is taken. The
image-plane metrics are unchanged between these two configurations, since both
derive \(\Delta u\), \(\Delta v\), and the LOS angle from the same unfiltered
box center. The depth filter then removes a further \(41\%\) of the residual
jitter, down to \(0.0264\,\mathrm m\), and the image-center filter improves
image-plane continuity on all three channels. The two stages are thus
complementary: the mask suppresses outliers from background disparities, and the
filters attenuate the residual measurement noise.

\subsection{Yaw-Control Ablation in Simulation}

A Unity simulation is used to isolate the contribution of the yaw channel: the
proposed model-fusion MPC is run with and without yaw control, and the errors
are reported as the norm of the 3D position error using MAE, P95, and maximum
error. In the ablated case the yaw reference is held at its initial value, so
that the two runs differ only in whether the vehicle is allowed to reorient
toward the target. The hydrodynamic damping is scaled to $30\times$ the engine
default so that the simulated vehicle exhibits the strongly damped, low-speed
response characteristic of the physical AUV, and the target traverses a
$0.75\,\mathrm{m}\times0.75\,\mathrm{m}$ rectangle at approximately
$0.15\,\mathrm{m/s}$.

Enabling yaw control reduces the MAE from $4.21$ to $3.22\,\mathrm{cm}$
($23.5\%$), with smaller gains in the P95 and maximum errors ($14.9\%$ and
$12.1\%$); Fig.~\ref{fig:rectangular-error} plots the corresponding
position-error norms. That the improvement is largest in the mean indicates that
yaw control acts mainly by shortening the recovery after each direction change
rather than by lowering the transient peak.

\subsection{Controller Comparison on Physical AUV}

\begin{figure}[!t]
    \centering
    \includegraphics[width=\linewidth]{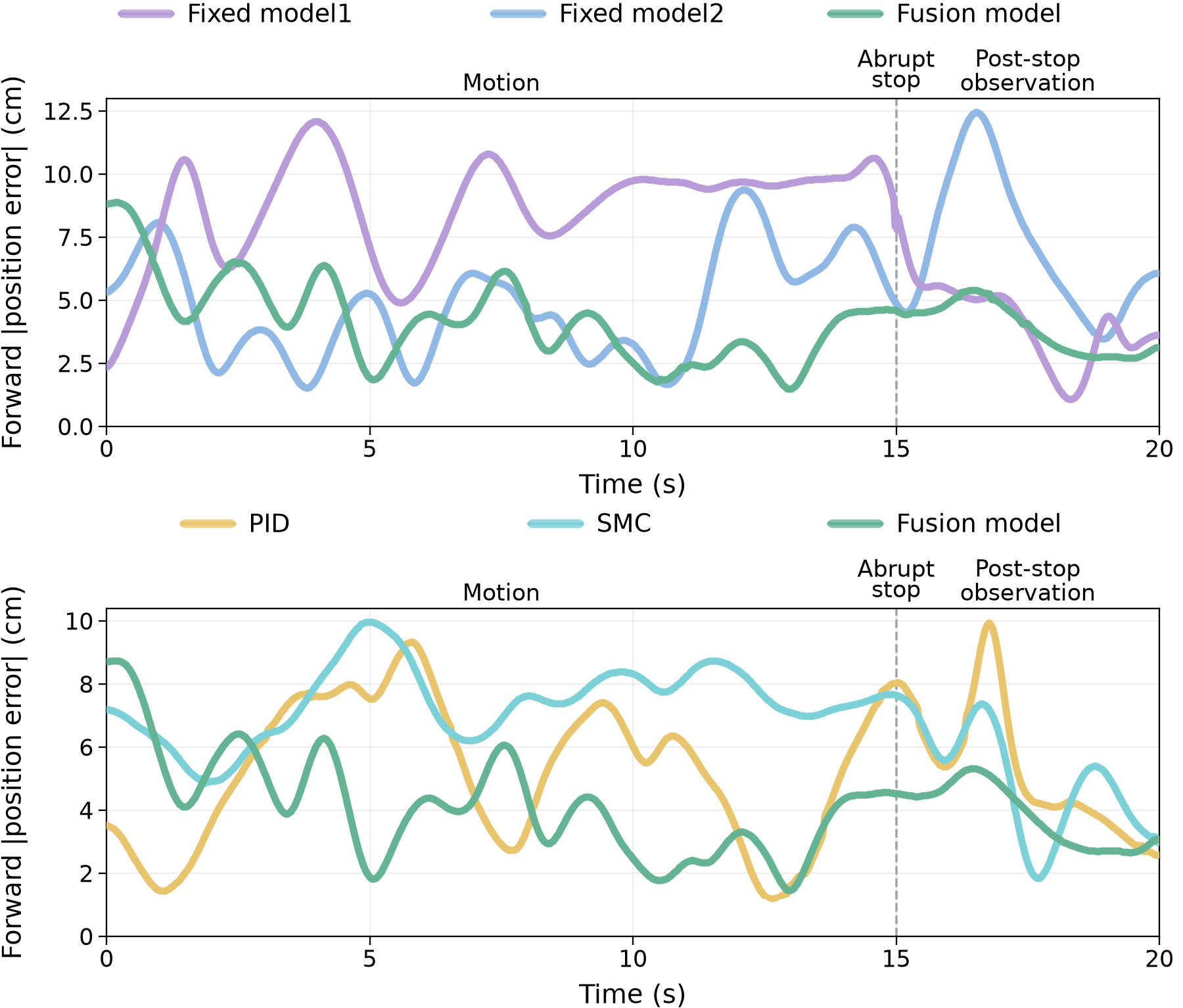}
    \caption{Absolute forward-position errors for the fusion and fixed-model controllers (top) and for PID, SMC, and the fusion model (bottom).}
    \Description{Absolute forward-position error over time, comparing the fusion
    controller against the two fixed-model variants (top) and against PID and
    SMC (bottom).}
    \label{fig:controller_comparison}
\end{figure}

\begin{table}[!t]
\caption{Forward-position errors over the common $20\,\mathrm{s}$ pool trial.}
\label{tab:pool}
\centering
\small
\setlength{\tabcolsep}{4pt}
\begin{tabular*}{\columnwidth}{@{\extracolsep{\fill}}lccc@{}}
\toprule
\textbf{Method} & \textbf{MAE}/\(\mathrm{cm}\,\downarrow\) &
\textbf{P95}/\(\mathrm{cm}\,\downarrow\) &
\textbf{Max}/\(\mathrm{cm}\,\downarrow\)\\
\midrule
PID & 5.23 & 11.81 & 20.59\\
SMC & 6.84 & 11.45 & \textbf{14.67}\\
Fusion model & \textbf{4.07} & \textbf{8.87} & 18.78\\
Fixed model 1 & 5.28 & 13.98 & 27.35\\
Fixed model 2 & 7.39 & 12.47 & 24.31\\
\bottomrule
\end{tabular*}
\end{table}

The controllers are compared on the physical AUV over a common $20\,\mathrm{s}$
pool trial, with errors reported along the forward ($X_b$) axis. The target
travels $200\,\mathrm{cm}$ during $15\,\mathrm{s}$ of acceleration and
constant-speed motion and then stops for $5\,\mathrm{s}$, so that the trial
exercises both the tracking and the settling behavior of each controller. All
methods share the relative-state estimation front-end described above and the
same reference standoff, so the comparison reflects the control stage alone.
Figure~\ref{fig:tracking-motion} shows the resulting tracking sequence, in which
the target stays near the image center while the vehicle follows it across the
pool.

Table~\ref{tab:pool} compares PID, SMC, the proposed fusion model, and its two
fixed-model variants in terms of MAE, P95, and maximum error. The fusion model
attains the lowest MAE and P95, reducing the MAE by $22.2\%$ and $40.5\%$
relative to PID and SMC and by $44.9\%$ and $22.9\%$ relative to the two
fixed-model variants. SMC attains the lowest maximum error, but its higher MAE
and P95 indicate a conservative response.

The top panel of Fig.~\ref{fig:controller_comparison} isolates the effect of the
model fusion by driving the same controller with Model~1 only (Fixed model~1),
Model~2 only (Fixed model~2), and their adaptive fusion. Each fixed model is
accurate only over the motion phase that matches its assumption. Fixed model~1
never predicts forward travel and therefore holds a large sustained error while
the target moves, whereas Fixed model~2 keeps predicting forward travel once the
target stops at $t=15\,\mathrm{s}$ and produces the largest post-stop excursion.
The fusion model stays near the lower envelope of both throughout the trial,
showing that the adaptive weights follow the transition from constant-speed
motion to rest instead of committing to either assumption. The bottom panel
compares the fusion model with PID and SMC, where it additionally damps the
overshoot that both exhibit at the stop.


\section{Conclusion and Future Work}

We presented an underwater visual-servoing framework for target tracking
with a multi-thruster AUV. Color--disparity--temporal mask fusion and decoupled filtering
improve relative-state stability, while adaptive model-fusion MPC handles
changing target motion under yaw and actuation constraints. Video replay,
simulation, and closed-loop pool experiments validate the complete system.
Future work will address disparity refinement, observation loss, and
open-water generalization.

\section*{Acknowledgment}

The authors used Claude Code and Codex for language polishing, data
organization, figure preparation, and code development. All AI-assisted
materials were carefully reviewed, verified, and approved by the authors, who take full
responsibility for the content and reported results.

\bibliographystyle{IEEEtran}
\bibliography{ref}

\end{document}